\documentclass[letterpaper,times]{IONconf}
\usepackage{amsmath}
\usepackage{bm}
\usepackage{amssymb}
\usepackage{algorithm}
\usepackage{algpseudocode}
\usepackage[sorting=none]{biblatex}
\bibliography{references}

\usepackage{tabularx}

\usepackage[pdftex]{graphicx}
\DeclareGraphicsExtensions{.pdf,.jpeg,.png}
\usepackage{url}
\title{Enhanced Laser-Scan Matching \protect\\ with Online Error Estimation\protect\\ for Highway and Tunnel Driving}

\author{
    Matthew McDermott, Jason Rife, \textit{Tufts~University}% <- this '%' removes a trailing whitespace
    }

\begin{document}

\maketitle
% ICET: laser scan matching, laser scan matching, managing ambiguities, online error estimation
% Adapting NDT to Estimate Error Covariance and Geometric Ambiguity in Laser Scan Matching
% Adapting Normal Distributions Transform to Estimate Error Covariance in Laser Scan Matching
% Improving Normal Distributions Transform for Highway and Tunnel Driving
%Enhanced Laser Scan Matching for Highway and Tunnel Driving

% best one so far
% Enhanced Laser-Scan Matching with Online Error Estimation for Highway and Tunnel Driving

% biography section. The * indicates a section excluded from numbering.
\section*{biography}

% Biographies are defined as follows:
% \biography{Author name}{author biography text}

\biography{Matthew McDermott}{is a student in the Mechanical Engineering Ph.D. program at Tufts University in Medford, MA. He works in the Automated Systems and Robotics Laboratory (ASAR) with Dr. Jason Rife. He received his B.S. and M.S. degrees in Mechanical Engineering at Tufts University.}

\biography{Jason Rife}{is a Professor and Chair of the Department of Mechanical Engineering at Tufts University in Medford, Massachusetts. He directs the Automated Systems and Robotics Laboratory (ASAR), which applies theory and experiment to characterize integrity of autonomous vehicle systems. He received his B.S. in Mechanical and Aerospace Engineering from Cornell University and his M.S. and Ph.D. degrees in Mechanical Engineering from Stanford University.
}

% The Abstract. The * indicates a section excluded from numbering.
\section*{Abstract}

Lidar data can be used to generate point clouds for the navigation of autonomous vehicles or mobile robotics platforms. Scan matching, the process of estimating the rigid transformation that best aligns two point clouds, is the basis for lidar odometry, a form of dead reckoning. Lidar odometry is particularly useful when absolute sensors, like GPS, are not available.
Here we propose the Iterative Closest Ellipsoidal Transform (ICET), a scan matching algorithm which provides two novel improvements over the current state-of-the-art Normal Distributions Transform (NDT). Like NDT, ICET decomposes lidar data into voxels and fits a Gaussian distribution to the points within each voxel. The first innovation of ICET reduces geometric ambiguity along large flat surfaces by suppressing the solution along those directions. The second innovation of ICET is to infer the output error covariance associated with the position and orientation transformation between successive point clouds; the error covariance is particularly useful when ICET is incorporated into a state-estimation routine such as an extended Kalman filter.
We constructed a simulation to compare the performance of ICET and NDT in 2D space both with and without geometric ambiguity and found that ICET produces superior estimates while accurately predicting solution accuracy.

% The introduction. Section numbering starts here.
\section{Introduction}

This paper introduces the Iterative Closest Ellipse Transform (ICET), a novel lidar scan matching algorithm. ICET is superior to current state-of-the-art methods of scan matching due to improvements in performance and interpretability. To the best of our knowledge, there is no existing algorithm that can, in addition to estimating transformation between point clouds, both produce online error estimates and predict ambiguous directions as a function of scene geometry. To understand the utility of the contributions of ICET, it is important to consider existing algorithms for lidar scan matching. 

One of the earliest scan matching methods is the Iterative Closest Point (ICP) algorithm \parencite{Besl}. ICP works by finding transformations that minimize the distances between corresponding points in two scans \parencite{Chen}.  Though simple, ICP  comes with a number of drawbacks, most importantly the challenge of  data association, where each point in the second scan must be matched correctly to a point in the first scan. Point-to-Plane ICP simplifies data association by fitting planes to groups of points and matching those planes between scans \parencite{Low}; however, the assumption of planar structure is not often representative of real-world scenes. 

The Normal Distribution Transform is an alternative to ICP and ICP Point-to-Plane that simplifies data association using an unstructured representation of the scene: a voxel grid \parencite{Biber}. The grid enforces spatial relationships among grid clusters without imposing specific geometric models to represent objects in the scene. NDT also replaces individual points with a density function thereby reducing sensitivity to noise, particularly during data association. The most simple implementation of NDT involves optimizing correspondences between distributions in the reference scan and individual points in the later scan (P2D-NDT), though more recent implementations have shown an increase in computational speed with similar performance by optimizing the correspondences between distribution centers of the two scans (D2D-NDT) \parencite{D2DNDT}. One common disadvantage of all variants of NDT, however, is  the lack of a quantitative indication of solution quality.

%TODO make it more clear that correspondences are in relation to CENTERS of distributions of first scan

%TODO add paragraph explaining all the parts of fig 1

\begin{figure}[h]
\centering
\includegraphics[width=5in]{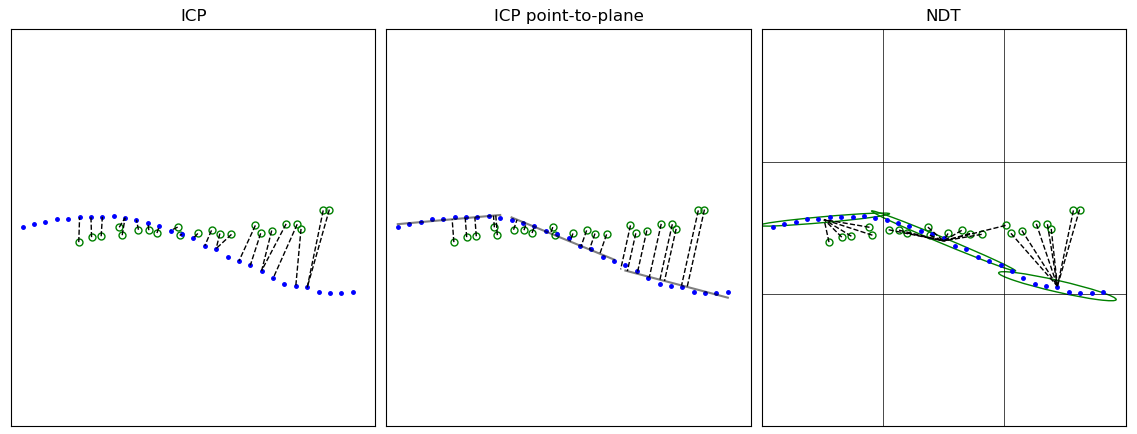}
% where an .eps filename suffix will be assumed under latex, 
% and a .pdf suffix will be assumed for pdflatex; or what has been declared
% via \DeclareGraphicsExtensions. Other suffixes like .png are added manually.
\caption{Visualization of ICP, ICP point-to-plane, and NDT. Dotted lines represent correspondences between various features in each scan.}
\label{fig:example}
\end{figure}

In many applications, machine-learning based approaches have been shown to be superior to traditional geometric analyses. With regards to scan matching, recent machine learning approaches such as LO-Net have achieved performance in dead reckoning similar to that of geometric based approaches like ICP and NDT \parencite{Li}. Unfortunately, the nature of end-to-end neural networks results in uninterpretable models where it is difficult to predict patterns of error. In the automotive application, it is particularly important to preserve some level of human readable information in order to develop a rigorous safety case.

Our new algorithm ICET addresses limitations of ICP, NDT and LO-Net. To overcome the data association issues of ICP, ICET uses a voxelized density function (much like NDT). To enable performance estimation, however, ICET leverages a nonlinear least-squares solution (as opposed to the optimization methods used in NDT). Error covariance matrices can be obtained from least-squares processing, but these covariance matrices are only made meaningful by introducing a new scene interpretation and dimension reduction step, a step not used in either ICP or NDT. The resulting algorithm enhances accuracy and characterizes errors; moreover, in contrast with LO-Net, ICET remains fully interpretable at every step.

The remainder of this paper presents the ICET algorithm and simulation results demonstrating its function. Section II highlights how geometric ambiguity arises and undermines existing scan matching methods. Section III describes our approach to scan matching using a linear least-squares formulation. Section IV introduces a technique to eliminate ambiguities from contributing to scan matching solution. Section V describes simulations to illustrate the algorithm's performance. Results of the simulation are discussed in Section VI. A final section summarizes the paper.

\section{Geometric Ambiguity}

\begin{figure}[H]
\centering
\includegraphics[width=2.5in]{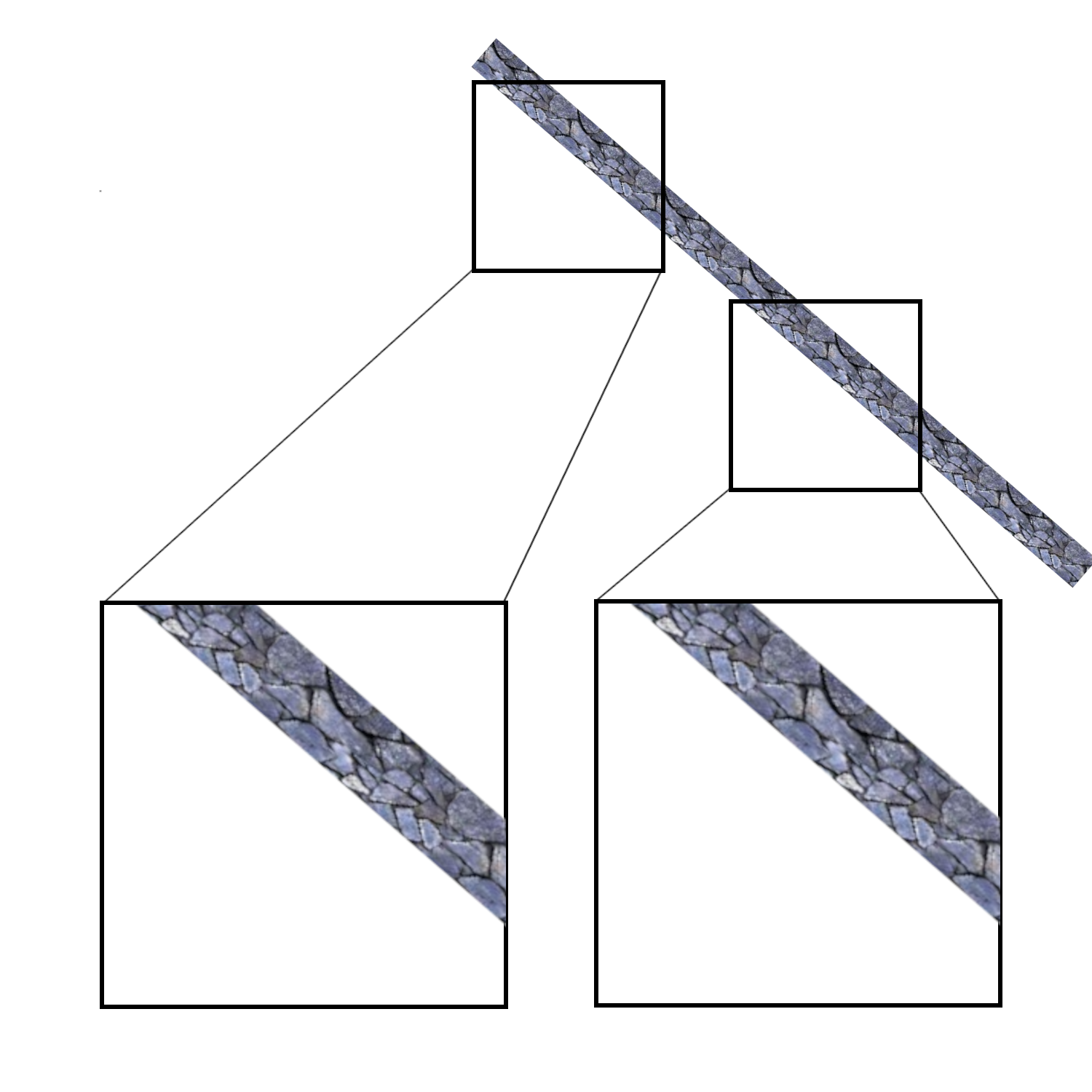}
\caption{Geometrically ambiguous extended surface }
\label{fig:ambiguity}
\end{figure}

The effectiveness of a scan matching technique is highly dependant on both the transformation between the two scans and the underlying geometry of the scene. While it is intuitive that too large a transformation between subsequent scans may make accurate estimation of their relationship impossible, it should be noted that the same is true even for small displacements in scenes that lack distinctive features in one or more direction. Figure \ref{fig:ambiguity} illustrates an example.

%TODO: add ref for each fig, make ref to fig1 in text

\begin{figure}[H]
\centering
\includegraphics[width=4in]{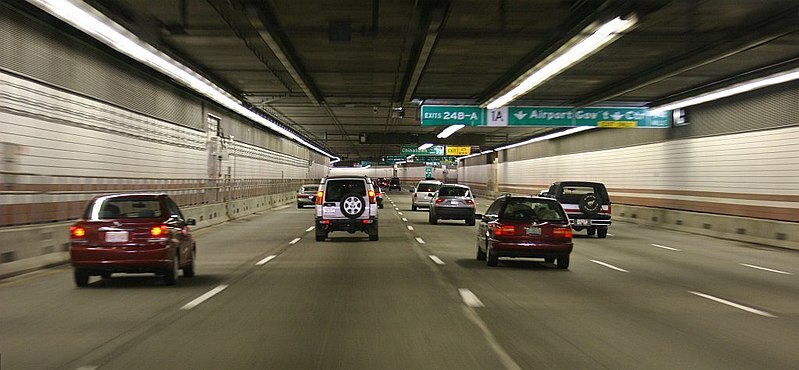} %was 3in for my photo
\caption{Road section with potential for high geometric ambiguity along the axis aligned with the road, courtesy of Rene Schwietzke
\parencite{tunnel}}
\label{fig:BostonAve}
\end{figure}

The figure visualizes an extended feature, which in this case is a wall. Two snapshots are taken at different locations along the wall.  A scan matching algorithm would most likely suggest that the two snapshots describe essentially the same physical location in the world, even though they are in fact drawn from distinct, non-overlapping locations. However, the same ambiguity would exist even if the scans were mildly overlapping, because the scans can only reliably be localized in the direction perpendicular to the wall and not along it. Similar scenes occur during driving on some highways, in  urban canyons, and in tunnels. In these environments, dominant terrain features (like the wall in Figure \ref{fig:ambiguity}) may be aligned with the road such that scan matching algorithms are only reliable perpendicular to the road and not along it. An example of an environment in which the dominant features are aligned with the road is shown in Figure \ref{fig:BostonAve}.

\begin{figure}[H]
\centering
\includegraphics[width=4.0in]{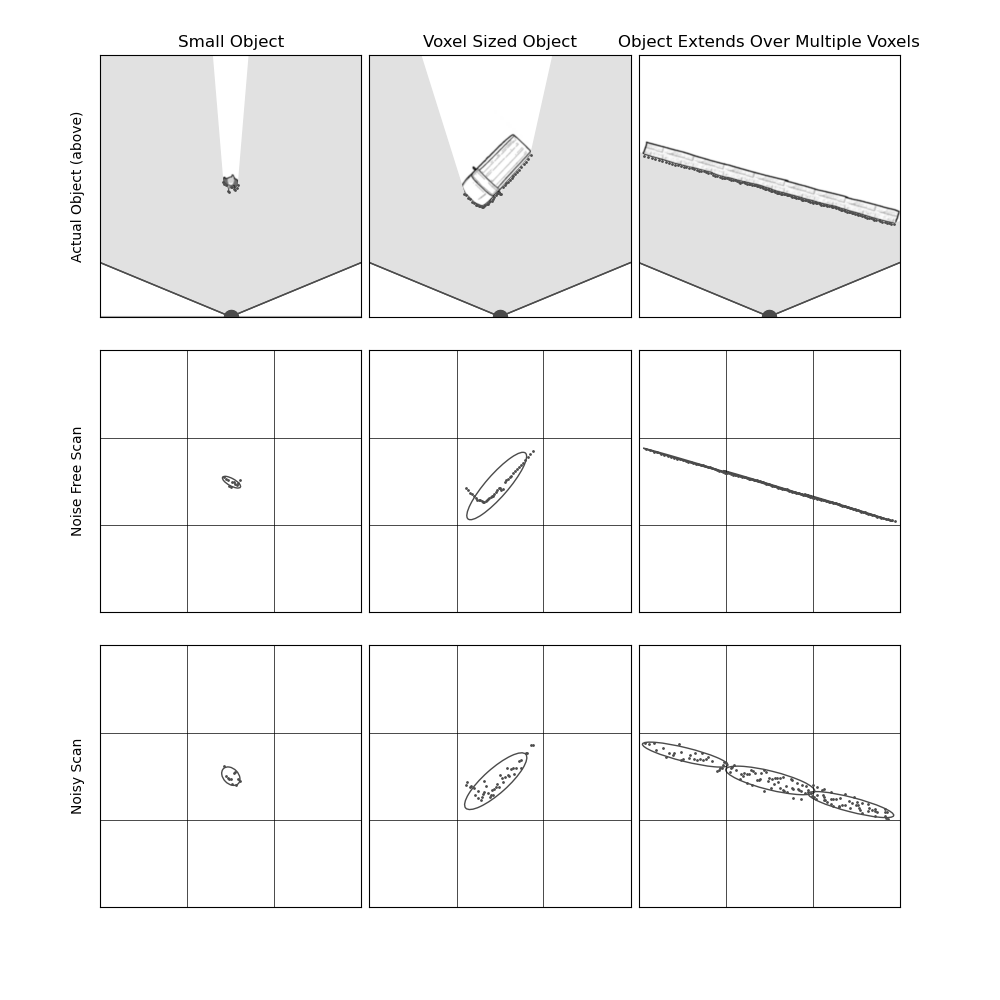}
% where an .eps filename suffix will be assumed under latex, 
% and a .pdf suffix will be assumed for pdflatex; or what has been declared
% via \DeclareGraphicsExtensions. Other suffixes like .png are added manually.
\caption{Voxelization of Lidar scans for three objects of different sizes: small, voxel-sized, and extended.}
\label{fig:gridExample}
\end{figure}

This ambiguity associated with extended objects, sometimes called the aperture problem~\cite{shimojo1989occlusion}, is  commonly  observed in vision-based and lidar-based localization. Although the aperture problem is well understood in some domains, its impact on lidar scan matching algorithms like NDT is not well understood. One of the primary goals of this paper is to introduce a new approach to mitigate the aperture problem when lidar scan points are voxelized as shown in Figure \ref{fig:gridExample}.

%Consider the situation in which a vehicle is traveling down a straight tunnel. Two subsequent lidar scans may show obvious rotation or cross-track displacement, but no information can possibly be gathered about along-track displacement. NDT, however, will still give a solution along the along-track direction, despite the fact that this displacement is not observable. Thus, NDT gives the false impression that unobservable states are computed reliably. One such geometrically ambiguous surface is visualized in Figure 2. Translation estimates from NDT along the direction of an extended feature are purely an artifact of the grid used in the voxelization process.

% Graphics are placed like in the following example (where the width should be adjusted depending on the image).

The flaw in the NDT algorithm is in its assumption that the lidar points within a voxel belong to a random distribution that lies entirely inside the voxel. This flaw is twofold in that the distribution is neither entirely random nor necessarily confined within a given voxel. These issues are visualized in Figure \ref{fig:gridExample}, which visualizes three objects (top row), simulated  lidar scans without noise (middle row), and lidar scans with simulated noise (last row). For a point-like object, one much smaller than the size of a grid voxel, the NDT assumptions are reasonable. In this case (left column of figure), the distribution of lidar samples is governed almost entirely by random sensor noise. For an object on the scale of a voxel (middle column), it becomes more clear that the distribution of scan points has a largely deterministic shape, a right angle shape for the vehicle visualized in the figure. In this middle case, NDT's assumption that the distribution is driven by purely random noise begins to break down; however, the Gaussian random-noise distribution assumed by NDT still provides a meaningful description of the object. When the object is extended (right column), however, NDT's assumptions break down completely. NDT represents the wall as multiple local Guassian distributions, each independent from its neighbors. Along the wall direction, localization errors larger than a voxel result in aliasing, with bad data association caused by lidar points  slipping into the next cell along the wall. This aliasing problem means that NDT implicitly creates an artificial upper bound for the noise along the length of the extended object. Because maximum distribution length is clipped by the length of a voxel, solutions from NDT are calculated with an implied standard deviation of error in the extended direction of less than one half of the voxel width.
This issue compounds the fact that no single error metric is generated by the NDT algorithm. In fact, without resolving the aperture problem, it is clear that it would be very difficult to extract a meaningful accuracy estimate directly from NDT. Our approach is to resolve these issues with a least-squares based approach that incorporates a mechanism for mitigating the aperture problem and that, thereby, enables meaningful online estimates of the accuracy of a lidar scan match.

\section{Scan Matching as a Linearized Least-Squares Problem}

This section re-envisions voxelized scan matching, replacing the optimization approach of NDT with a least-squares solution in order to introduce an analytical approach for estimating solution accuracy. Obtaining an accuracy estimate for a voxelized lidar scan match is the first novel feature of our ICET algorithm. 

ICET  works by estimating the solution vector \textbf{x} that best represents the transformation between two point clouds.  This paper will focus on the 2D case, where \textbf{x} consists of one rotation angle and two scalar translations: $\theta$, $x$ and $y$. 

\begin{equation}\label{eq:xvec}
    \textbf{x} = \begin{bmatrix}
        x \\
        y \\
        \theta \\
    \end{bmatrix}
\end{equation}

For the 3D case, by comparison, the vector would consist of a generalized rotation description (e.g. Euler angles or quaternions) and a 3D translation vector. In both the 2D and 3D cases, the transformation vector $\mathbf{x}$ maps the points in the second (or \textit{new}) scan into the coordinate system used by the first (or \textit{reference}) scan. We assign 2D positions $^{(i)}\textbf{p}$ to describe the location of each point $i$ in the new scan using Cartesian \{x,y\} coordinates defined in the body-fixed frame at the time of the new scan. The goal of the scan match is to transform the coordinates $^{(i)}\textbf{p}$ into an alternative set of coordinates $^{(i)}\textbf{q}$, corresponding to the original body-fixed frame (at the time reference scan). The $^{(i)}\textbf{q}$ are related to the states in (\ref{eq:xvec}) as described below.

%% NOTE >>> VARIABLE DEFINITION CHANGE >> I intentionally flipped the p and q below (compared to the old PDF) and redefined q to describe original coordinate system, while p now means new coordinate system

\begin{equation}\label{eq:twoDtransform}
        ^{(i)}\textbf{q}(\textbf{x}) = \begin{bmatrix}
                cos(\theta) & -sin(\theta) \\
                sin(\theta) & cos(\theta) \\
        \end{bmatrix}
         {}^{(i)}\textbf{p} - 
        \begin{bmatrix}
                x \\
                y \\
        \end{bmatrix}
\end{equation}

Averaging the $^{(i)}\textbf{q}(\textbf{x})$ over a voxel gives the voxel mean. Concatenating all of those voxel means into a larger vector gives a nonlinear observer function $\textbf{y}=\textbf{h}(\textbf{x})$, which can be compared to a similar concatenated voxel-mean vector $\mathbf{y}_0$ from the original image. Note that here, and subsequently in this paper, variable pairs like $\mathbf{y}$ and $\mathbf{y}_0$ are defined such that the case with no subscript indicates the new scan and the case with the ``$0$'' subscript, the reference scan. Ideally, the voxel-mean vectors are identical for both scans ($\mathbf{y} = \mathbf{y}_0$), so  $\mathbf{x}$ can be estimated by inverting the observation equation to solve

\begin{equation}\label{eq:nonlinEq}
   \mathbf{h}(\mathbf{x}) = \mathbf{y}_0.
\end{equation}

The goal of ICET is to solve (\ref{eq:nonlinEq}) using an iterative Newton-Raphson approach. The approach works by linearizing the nonlinear equation via a Taylor series expansion. 

\begin{equation}
    \textbf{y}_0 = \textbf{h}(\hat{\textbf{x}}) + \textbf{H}\delta \textbf{x} + \mathcal{O}(\delta \textbf{x}^2)
\end{equation}

Here \textbf{H} is the \textit{Jacobian}, a matrix of first derivatives of the function $\textbf{h}$ with respect to each element of $\textbf{x}$. In the Taylor Series expansion, the current best estimate is $\hat{\textbf{x}}$. A linear correction $\delta \textbf{x}$ can be computed assuming that the higher-order terms are zero:

\begin{equation}\label{eq:linearizedDeltaY}
    \textbf{H} \delta \textbf{x} = \Delta \textbf{y}
\end{equation}

Here $\Delta \textbf{y}=\textbf{y}_0-\textbf{h}(\hat{\textbf{x}})$. Introducing a weighting matrix $\textbf{W}$, the standard weighted least-squares solution~\cite{Simon} is 

\begin{equation}\label{eq:dxCorrection}
    \delta \textbf{x} = (\textbf{H}^T\textbf{WH})^{-1} \textbf{H}^T\textbf{W} \Delta \textbf{y}.
\end{equation}

After computing $\delta \textbf{x}$ using (\ref{eq:dxCorrection}), we can update the estimate $\hat{\textbf{x}}$ according to the following equation, and iterate to convergence.

\begin{equation}\label{eq:replacement}
    \hat{\textbf{x}} \xrightarrow[]{} \hat{\textbf{x}} + \delta \textbf{x}
\end{equation}

In order to implement this approach, it is necessary to specify certain terms used above, by providing the precise definitions of the voxel means  $\textbf{y}$  and $\textbf{y}_0$, the Jacobian $\textbf{H}$, and the weighting matrix $\textbf{W}$. First consider the voxel means. A subset of points $i$ from a scan belong to a given voxel $j$. For the original scan  these are $i \in ^{(j)}\mathcal{I}_0$ and for the new scan, $i \in ^{(j)}\mathcal{I}$.  For a given voxel, the mean $^{(j)}\textbf{y}_0$ of the new-scan points and the mean $^{(j)}\textbf{y}$ of the reference-scan points are computed as follows.

\begin{equation}\label{eq:voxelY0}
    ^{(j)}\textbf{y}_0 = \frac{1}{|^{(j)}\mathcal{I}_0|} {\sum_{i \in ^{(j)}\mathcal{I}_0}} ^{(i)}\mathbf{q_0}
\end{equation}

\begin{equation}\label{eq:voxelY}
    ^{(j)}\textbf{y}(\hat{\textbf{x}}) = 
    \frac{1}{|^{(j)}\mathcal{I}|} {\sum_{i \in ^{(j)}\mathcal{I}}} ^{(i)}\mathbf{q(\hat{\textbf{x}})}
\end{equation}

The number of points in each subset is the cardinality of that subset, indicated by the ``$|*|$'' notation. Statistics are only computed for those voxels where the number of lidar points exceeds a minimum cutoff; correspondences are then generated for voxels, such that there remains a subset of voxels, indexed  $j \in J$, matched between the new and reference scan. The corresponding means (for voxels  $j \in J$) are concatenated to form the larger observation vectors: $\mathbf{y}_0$ and $\mathbf{y}=\mathbf{h}(\hat{\textbf{x}})$.

\begin{equation}
    \mathbf{y}_0 = \begin{bmatrix}
                ^{(1)}\textbf{y}_0 \\
                ^{(2)}\textbf{y}_0 \\
                ...\\
                % .\\
                % .\\
                ^{(J)}\textbf{y}_0\\
                \end{bmatrix}
\end{equation}

\begin{equation}\label{eq:fullY}
    \textbf{h}(\hat{\textbf{x}}) = \begin{bmatrix}
                ^{(1)}\textbf{y}(\hat{\textbf{x}}) \\
                ^{(2)}\textbf{y}(\hat{\textbf{x}}) \\
                ...\\
                ^{(J)}\textbf{y}(\hat{\textbf{x}})\\
                \end{bmatrix}
\end{equation}

Computing the $\delta\mathbf{x}$ correction in (\ref{eq:dxCorrection}) requires differencing the above two vectors to give $\Delta\textbf{y}=\textbf{y}_0-\textbf{h}(\hat{\textbf{x}})$. If the dimension of the scan is $N$, with $N=2$ in the 2D case and $N=3$ in the 3D case, then $\Delta\textbf{y} \in \mathbb{R}^{JN}$. Taking the derivative of (\ref{eq:fullY}) with respect to each variable in $\mathbf{x}$, we can write the Jacobian $\textbf{H}$ in terms of a submatrix $^{(j)}\textbf{H}$ for each $j \in J$.

\begin{equation}
    \textbf{H} = \begin{bmatrix}
    ^{(1)}\textbf{H} \\
    ... \\
    ^{(J)}\textbf{H} \\
    \end{bmatrix}
\end{equation}

For the 2D case, the  $^{(j)}\textbf{H}$ are obtained by substituting (\ref{eq:twoDtransform}) into (\ref{eq:voxelY}) and taking derivatives with respect to each state in (\ref{eq:xvec}).

\begin{equation}
    ^{(j)}\textbf{H} = 
  \left( 
    \begin{array}{cc|c}
        -1 & 0 & ^{(j)} \mathbf{H}_{\theta} \\
        0 & -1  \\
    \end{array}
    \right)
\end{equation}

Here the vector ${}^{(j)}\mathbf{H}_{\theta}$ is computed by evaluating derivatives of (\ref{eq:nonlinEq}) with respect to $\theta$ and summing over $i \in {}^{(j)}\mathcal{I}$.

\begin{equation}
    ^{(j)} \mathbf{H}_{\theta} = \frac{1}{|^{(j)}\mathcal{I}|} \sum_{i \in {}^{(j)}\mathcal{I}} \begin{bmatrix}
    -sin(\theta) & -cos(\theta) \\
    cos(\theta) & -sin(\theta) \\
    \end{bmatrix}  {}^{(i)}\textbf{p}
\end{equation}

%% Resume Here %

To evaluate the $\delta\mathbf{x}$ correction in (\ref{eq:dxCorrection}), we must still define the weighting matrix $\mathbf{W}$. For an optimal solution~\cite{Simon}, $\Delta\mathbf{y}$ should be unbiased and the covariance of $\Delta\mathbf{y}$ should be inverted to form the weighting matrix $\mathbf{W}$. The covariance of the observation vector is typically called the \textit{sensor-noise covariance} and labeled $\mathbf{R}$. Thus we expect the weighting matrix to have this form:

\begin{equation}
    \textbf{W} = \textbf{R}^{-1}
\end{equation}

In many navigation applications, the matrix $\mathbf{R}$ is determined strictly from a model of sensor noise. In the case of scan matching, we have a luxury in that each voxel mean is computed statistically from many data points, and so we can directly estimate the covariance from the data.  For a given voxel $j$, define the reference-scan covariance estimate (computed statistically as the central second moment over the point locations) to be $^{(j)}\textbf{Q}_0$, and define the new-scan estimate to be $^{(j)}\textbf{Q}$.  

\begin{equation}\label{eq:covQ0}
    {}^{(j)}\textbf{Q}_0 = \frac{1}{
|{}^{(j)}\mathcal{I}_0|-1} {\sum_{i \in {}^{(j)}\mathcal{I}_0}} \Big({}^{(i)}\mathbf{q}_0-\mathbf{y}_0\Big)\Big(
{}^{(i)}\mathbf{q}_0-\mathbf{y}_0\Big)^T
\end{equation}

\begin{equation}\label{eq:covQ}
    {}^{(j)}\textbf{Q}(\hat{\mathbf{x}}) = \frac{1}{
|{}^{(j)}\mathcal{I}|-1} {\sum_{i \in {}^{(j)}\mathcal{I}}} \Big({}^{(i)}\mathbf{q}(\hat{\mathbf{x}})-\mathbf{y}\Big)\Big(
{}^{(i)}\mathbf{q}(\hat{\mathbf{x}})-\mathbf{y}\Big)^T
\end{equation}

A well-known result is that the covariance ${}^{(j)}\mathbf{\Sigma}_{y_0}$ of the sample mean vector ${}^{(j)}\mathbf{y}_0$ can be related to the true covariance, using (\ref{eq:voxelY0}) and the central limit theorem~\parencite{NIST}. Given a sufficient number of samples in a voxel, the estimated covariances of (\ref{eq:covQ0}) and (\ref{eq:covQ}) can be substituted for the true covariances. Taking this approach, we estimate ${}^{(j)}\mathbf{\Sigma}_{y_0}={}^{(j)}\mathbf{Q}_0 / |{}^{(j)}\mathcal{I}_0|$ and ${}^{(j)}\mathbf{\Sigma}_{y}={}^{(j)}\mathbf{Q} / |{}^{(j)}\mathcal{I}|$. 

Next, we note $\mathbf{R}$ describes the variance of  $\Delta\mathbf{y}$, which is formed by differencing and concatenating the mean vectors ${}^{(j)}\mathbf{y}$ and ${}^{(j)}\mathbf{y}_0$ for all voxels $j \in J$. For each voxel, a local covariance ${}^{(j)}\mathbf{R}$ can be defined considering contributions from ${}^{(j)}\mathbf{y}$ and ${}^{(j)}\mathbf{y}_0$. In other words, ${}^{(j)}\mathbf{R} = {}^{(j)}\mathbf{\Sigma}_{y_0}+{}^{(j)}\mathbf{\Sigma}_{y}$. Assuming point distributions are uncorrelated across voxels, then the full covariance matrix $\mathbf{R}$ is block diagonal with diagonal elements ${}^{(j)}\mathbf{R}$. That is:

\begin{equation}
    \mathbf{R} = \begin{bmatrix}
        {}^{(1)}\mathbf{R} & \mathbf{0} & \dots & \mathbf{0}  \\
        \mathbf{0} & {}^{(2)}\mathbf{R} & \dots &  \mathbf{0} \\
        \vdots & \vdots & \ddots & \vdots
        \\
        \mathbf{0} & \mathbf{0} & \dots & {}^{(J)}\mathbf{R} 
    \end{bmatrix} 
\end{equation}

When the number of voxels is large, the covariance matrix $\mathbf{R}$ is rather sparse, so it is inefficient to construct the full matrix. Instead, a more efficient mechanism for evaluating  (\ref{eq:dxCorrection}) is to recognize that

\begin{equation}\label{eq:condense1}
    \mathbf{H}^T\mathbf{WH} = {\sum_{j \in J}} \Bigg[{}^{(j)}\mathbf{H}^T\,\ {}^{(j)}\mathbf{R}^{-1}\,\ 
    {}^{(j)}\mathbf{H}\Bigg]
\end{equation}

and

\begin{equation}\label{eq:condense2}
     \textbf{H}^T\textbf{W} \Delta \textbf{y} = {\sum_{j \in J}} \Bigg[{}^{(j)}\mathbf{H}^T\,\  {}^{(j)}\mathbf{R}^{-1}\,\ 
    \Big({}^{(j)}\mathbf{y}_0-{}^{(j)}\mathbf{y}(\hat{\mathbf{x}})\Big) \Bigg].
\end{equation}

Substituting (\ref{eq:condense1}) and (\ref{eq:condense2}) into (\ref{eq:dxCorrection}) greatly increases efficiency for computing the correction $\delta\mathbf{x}$. Equation (\ref{eq:condense1}) is also very useful for computing the covariance $\mathbf{P}$, which describes the error of the estimated state $\hat{\mathbf{x}}$ relative to the true state $\mathbf{x}_{true}$. 

\begin{equation}
    \mathbf{P}= E[(\hat{\mathbf{x}}-\mathbf{x}_{true})(\hat{\mathbf{x}}-\mathbf{x}_{true})^T]
\end{equation}

Assuming errors are small perturbations around the solution, it is well known \cite{Simon} that the linearized equations result in the following expression for the covariance.

\begin{equation}\label{eq:Pmatrix}
    \mathbf{P} = (\mathbf{H}^T\mathbf{WH})^{-1}
\end{equation}

This last equation represents ICET's  prediction of its own accuracy, a prediction that automatically accounts for the geometric distribution of useful voxels and for the quality of the scan points in those voxels.

\section{Ambiguity Management via Dimension Reduction}\label{AmbiguitySec}
One limitation of the prior section is that the methods assume the lidar points within each voxel are scattered about the mean due to purely random sensor noise. In fact, objects and natural terrain shape the point distribution, such that the observed scan has some degree of deterministic structure. The weighted least-squares solution is optimal only if the point distribution is random,  so this deterministic structure degrades solution accuracy.

Our approach to limiting the impact of unmodeled deterministic structure is to exclude measurements in directions where the covariance matrix approaches the size of the voxel. Deterministic structure is particularly problematic for objects that extend across multiple voxels (e.g. a long wall), because the within-voxel covariance provides a misleadingly optimistic estimate of measurement accuracy. To detect large variance values and identify their direction, we perform an eigendecomposition on the reference-scan covariance matrix ${}^{(j)}\mathbf{Q}_0$ for each voxel $j \in J$. Each ${}^{(j)}\textbf{Q}_0$ can be coverted to an eigenvalue matrix ${}^{(j)}\mathbf{\Lambda}$ and an eigenvector matrix ${}^{(j)}\mathbf{U}$ and decomposed as follows.

\begin{equation}\label{eq:eigen1}
    {}^{(j)}\mathbf{Q}_0 = {}^{(j)}\mathbf{U} \,\,\, {}^{(j)}\mathbf{\Lambda}\,\,\, {}^{(j)}\mathbf{U}^T
\end{equation}

The eigenvalue matrix describes the principal axis lengths (squared) for the variance ellipse. The eigenvector matrix describes the directions of those principal axes.  By testing each eigenvalue to see if it exceeds a reasonable threshold, we can identify overly extended distributions.
The eigenvalue threshold $T$ is based on the voxel width $a$. It is well known that extended objects of uniform density along their length (e.g. a uniform-density bar) have a variance in that direction of $a^2/12$. This result is tabulated in introductory mechanics textbooks like~\cite{beer2007mechanics}, where the variance, or central second moment, is called a \textit{moment of inertia}. Noting covariance matrices computed from lidar data are stochastic, the threshold is reduced slightly to avoid missed detection of extended objects, to a value of $T=a^2/16$. We also restrict the minimum voxel width to be significantly wider than the standard deviation of the lidar noise, so that the false detection risk is low for small objects and normal to surfaces.

Without loss of generality, we can partition the eigenvalue matrix into two blocks, a first block $\mathbf{\Lambda}_P$ that includes the values less than the threshold $T$ and a second block $\mathbf{\Lambda}_N$ that includes the values greater than or equal to the threshold.

\begin{equation}
    \mathbf{\Lambda} = \begin{bmatrix}
        \mathbf{\Lambda}_P & 0 \\
        0 & \mathbf{\Lambda}_N \\
    \end{bmatrix}
\end{equation}

Similarly, we can partition the corresponding eigenvectors into two submatrices, the eigenvectors $\mathbf{U}_P$ describing the dimensions preserved and the eigenvectors $\mathbf{U}_N$ describing the dimensions eliminated.

\begin{equation}
    \mathbf{U} = \begin{bmatrix}
        \mathbf{U}_P & \mathbf{U}_N \\
    \end{bmatrix}
\end{equation}

We can then project each voxel-mean vector into the preserved directions, eliminating the other directions. This operation is only necessary when the eigenvalue dimension is reduced yet not trivial (i.e. when $\mathbf{\Lambda} \in \mathbb{R}^{N \times N}$, $\mathbf{\Lambda}_P \in \mathbb{R}^{n \times n}$, and $0<n<N$). To account for all cases, we introduce a modified voxel mean, indicated by a tilde, as in  ${}^{(j)}\tilde{\mathbf{y}}$.

\begin{equation}\label{eq:ytilde}
    {}^{(j)}\tilde{\mathbf{y}} = \begin{cases}
    {}^{(j)}\mathbf{y}, &  n=N \\
    {}^{(j)}\mathbf{U}_P^T\,\ {}^{(j)}\mathbf{y}, & 0<n<N \\
    \emptyset, & n=0
  \end{cases}
\end{equation}

The third instance of (\ref{eq:ytilde}), the null result, occurs only for distributions wide in all directions, as in the case of loose foliage. 

To compute our solution via (\ref{eq:dxCorrection}), the dimension-reduction process must also be applied to other variables including ${}^{(j)}\mathbf{y}_0$, ${}^{(j)}\mathbf{H}$ and ${}^{(j)}\mathbf{R}$.
First, the voxel means ${}^{(j)}\mathbf{y}_0$ for the reference scan become ${}^{(j)}\tilde{\mathbf{y}}_0$, which has a structure analagous to (\ref{eq:ytilde}), but with ${}^{(j)}\mathbf{y}_0$ substituted for ${}^{(j)}\mathbf{y}$. Second, reducing the dimensions of ${}^{(j)}\mathbf{H}$ and ${}^{(j)}\mathbf{R}$ gives ${}^{(j)}\tilde{\mathbf{H}}$ and ${}^{(j)}\tilde{\mathbf{R}}$, defined as follows.

\begin{equation}\label{eq:Htilde}
    {}^{(j)}\tilde{\mathbf{H}} = \begin{cases}
    {}^{(j)}\mathbf{H}, &  n=N \\
    {}^{(j)}\mathbf{U}_P^T\,\ {}^{(j)}\mathbf{H}, & 0<n<N \\
    \emptyset, & n=0
  \end{cases}
\end{equation}

\begin{equation}\label{eq:Rtilde}
    {}^{(j)}\tilde{\mathbf{R}} = \begin{cases}
    {}^{(j)}\mathbf{R}, &  n=N \\
    {}^{(j)}\mathbf{U}_P^T\,\ {}^{(j)}\mathbf{R} \,\ {}^{(j)}\mathbf{U}, & 0<n<N \\
    \emptyset, & n=0
  \end{cases}
\end{equation}

In all instances, the \textit{tilde} indicates the dimension-reduction process. Using the dimension-reduced variables, solution steps (\ref{eq:condense1}) and (\ref{eq:condense2}) can be rewritten, respectively, as follows.

\begin{equation}\label{eq:condense3}
    \mathbf{H}^T\mathbf{WH} = {\sum_{j \in J}} \Bigg[{}^{(j)}\tilde{\mathbf{H}}^T\,\  {}^{(j)}\tilde{\mathbf{R}}^{-1}\,\ 
    {}^{(j)}\tilde{\mathbf{H}}\Bigg]
\end{equation}

\begin{equation}\label{eq:condense4}
     \textbf{H}^T\textbf{W} \Delta \textbf{y} = {\sum_{j \in J}} \Bigg[{}^{(j)}\tilde{\mathbf{H}}^T\,\  {}^{(j)}\tilde{\mathbf{R}}^{-1}\,\ 
    \Big({}^{(j)}\tilde{\mathbf{y}}_0-{}^{(j)}\tilde{\mathbf{y}}\Big) \Bigg].
\end{equation}

Equations (\ref{eq:condense3}) and (\ref{eq:condense4}) can be substituted into (\ref{eq:dxCorrection}) and, in most cases, iterated to obtain the Newton-Raphson solution.  There is one corner case to consider, however.  If all of the information in one direction is removed (e.g. for the case where the only feature is a straight wall that stretches across the entire scan), then (\ref{eq:condense3}) will be rank deficient and therefore non-invertible. A method for handling this special case is discussed in the Appendix.

The full algorithm, including both the basic solution and dimension reduction, is summarized in the table labeled Algorithm~\ref{tab:algo}.

\begin{algorithm}
    \begin{algorithmic}[1]
    \caption{ICET}
    \label{tab:algo}
    \State Initialize solution vector \textbf{x} \Comment{(1)}
    \State Break down reference scan into voxels
    \State Calculate mean and covariance matrix for voxels in reference scan with more than $\tau$ points \Comment{(\ref{eq:voxelY0}), (\ref{eq:covQ0})}
    \State Calculate $U$ and $L$ matrices for the reference scan in each qualifying voxel \Comment{(\ref{eq:eigen1})}
    
    \While{ $\delta\textbf{x}$ is converging}
    \State Transform second point cloud by \textbf{x} \Comment{(2)}
    
    \State Voxelize transformed point cloud using the same grid used to subdivide the reference scan
    
    \State Compute the new-scan mean and covariance of lidar points in each qualifying voxel \Comment{(\ref{eq:voxelY}),(\ref{eq:covQ})}
    
    %was this
    % \State Define a correspondence for each voxel qualifying for both scans
    
    \State Define a nearest-neighbor correspondence between distributions in each scan
    
    % \State For each element of $y$:
    
    %TODO: j was used earlier to describe each usable voxel, this is effectively the same thing, so leave as is?   
    \For{each correspondence \textit{j}}
    
        \State Compute the reduced-dimension means ${}^{(j)}\tilde{\mathbf{y}}_0$ and ${}^{(j)}\tilde{\mathbf{y}}$\Comment{(\ref{eq:ytilde})}

        \State Calculate ${}^{(j)}\tilde{\mathbf{H}}$ \Comment{(\ref{eq:Htilde})}
        
        \State Calculate the sensor-noise matrix ${}^{(j)}\tilde{\mathbf{R}}$\Comment{(\ref{eq:Rtilde})}
        
        \State Compute intermediate quantities as running sums  \Comment{(\ref{eq:condense3}),(\ref{eq:condense4})}
        
    \EndFor

    \State Calculate condition number of $\mathbf{H}^T \mathbf{W H}$ \Comment{(\ref{eq:eigen2})}
    
    \If {condition number below cutoff} 
    
        \State Determine  the state correction $\delta \mathbf{x}$  \Comment{(\ref{eq:dxCorrection})}
    
        \State $\hat{\mathbf{x}} \xrightarrow{} \hat{\mathbf{x}} + \delta \mathbf{x}$ \Comment{(\ref{eq:replacement})}
        
        \State Calculate state-error covariance $\mathbf{P}$ \Comment{(\ref{eq:Pmatrix})}
    
    \Else {  (See Appendix for details)}
    
        \State Determine  the subspace correction $\delta \mathbf{z}$ and update $\mathbf{x}'$  \Comment{(\ref{eq:dzCorrection}),(\ref{eq:replacement2})}
    
        \State $\hat{\mathbf{x}} \xrightarrow{} \hat{\mathbf{x}} + \mathbf{x}'$\Comment{(\ref{eq:update2})}
        
        %TODO how does this differ from just ignoring unused cov elements from P?
        \State Calculate subspace-error covariance $\mathbf{\Gamma}_P^{-1}$ \Comment{(\ref{eq:reducedCov})}
    
    \EndIf
    
\EndWhile
\end{algorithmic}
\end{algorithm}

\section{Simulation}

\begin{figure}[H]
\centering
\includegraphics[width=3.025in]{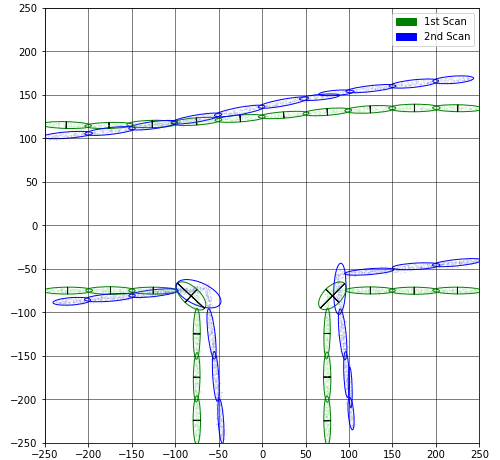} %was 2in
\includegraphics[width=3.0in]{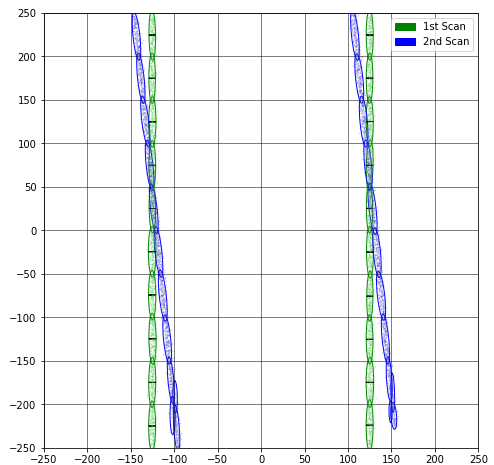}
\caption{Sample transformations for T-intersection and straight tunnel case. Voxels shown as square grid cells. Each ellipse represents the two-sigma covariance ellipse for a grid cell. Lines are drawn across dimensions of distributions in the first scan that have not been reduced.}
\label{fig:twoRoads}
\end{figure}

To demonstrate the central contributions of this paper, ICET was benchmarked against NDT. An implementation of NDT was programmed from scratch using the standard algorithm and optimization routine provided by Biber~\cite{Biber}. Monte-Carlo (MC) simulations of NDT and ICET were conducted through two scenarios featuring simple, simulated driving environments, as illustrated in Figure~\ref{fig:twoRoads}. The first scenario represents a T-intersection. The second scenario represents a straight tunnel or roadway. The T-intersection case involves a scene in which there are both along-track and cross-track features present in the frame. The straight-tunnel case involves a scene of high geometric ambiguity in which there are only along-track features. 
Both cases were represented in two-dimensions, with vertical walls assumed on the sides of the roadway. The units of the environment were somewhat arbitrary (but each unit length is meant to correspond to approximately 0.05 cm of physical distance). 

Each MC simulation considered only one environment. For a given environment, each MC trial considered a pair of scans: a reference scan and a new scan depicting the scene after movement.  Examples of scan pairs are shown in Figure~\ref{fig:twoRoads}, where the reference scan is green and the new scan, blue. Grid lines in the figure correspond to voxel boundaries, with voxels of about 50 units on a side (about 2.5 m). Mean and covariance of lidar points in each voxel is ilustrated by a two-sigma ellipse.  The illustrations indicate that the motion was small, with a translation of about 5 units laterally and 10 units vertically (about 0.25 m laterally and 0.5 m vertically). The yaw rotation between scans was 0.1 radian. The same translation and rotation was applied between scans in all trials, but lidar noise was re-sampled for each trial. Simulated lidar scans were constructed with 4200 samples. Gaussian noise with a standard deviation of 2 units in the $x$ and $y$ directions was applied to each sample. 

Note that both the ICET and NDT algorithms were implemented with two correspondence methods.  The baseline approach  defines correspondences only for means co-located within the same grid cell. A modified approach defines correspondences using the \textit{1-NN} or "nearest neighbor" method, which provides more robust initialization. Simulations were run using both correspondence methods with no statistically significant difference in solution accuracy. Therefore, only results from simulations using the nearest-neighbor approach are presented below. 

\section{Results and Discussion}

Results were computed from MC simulations with 1000 trials for each environment. In each trial, the error was computed as $\hat{\mathbf{x}}-\mathbf{x}_{true}$.  Mean and variance for lateral (or $x$-direction) error, longitudinal (or $y$-direction) error, and rotational errors were computed for each environment, for both the NDT and ICET algorithms. A predicted state-error covariance matrix was also computed for ICET.

The distributions appeared zero centered, with no evidence of a systematic mean error across trials. Standard deviations (labeled std error) are tabulated below (in simulation units) for the two translation directions and for rotation. 

The tunnel geometry was specially configured to suppress all geometric information in one dimension, in order to create a singularity in the computation of (\ref{eq:dxCorrection}). To address this issue, a specialty solution was applied, as defined by (\ref{eq:dzCorrection}) in the Appendix. In all of the tunnel-geometry trials, ICET was able to correctly identify the direction of geometric ambiguity and suppress the solution along that axis. For this reason the $y$ direction error is listed as N/A for ICET; the measurements were identified as ambiguous, so no solution was attempted in that direction. 

The error standard deviations for ICET were consistently better by nearly an order of magnitude as compared to NDT, presumably because the dimension reduction process eliminates processing in ambiguous directions, which NDT believes to be well-characterized (e.g. modeling point-cloud uncertainty based on the voxel dimension, even for environmental features that extend across many voxels). The extreme case of this trend occurred for the tunnel environment, where NDT delivered a lateral error two orders of magnitude larger than the longitudinal error, without enunciating any warning about a potential fault caused by ambiguity associated with extended wall features.

The predictive error-covariance matrices computed internally by the ICET algorithm (labeled as "ICET Predicted") were consistently representative of the true error distribution (labeled "ICET Actual").

\subsection*{T-Intersection Case}

 \begin{tabularx}{1.0\textwidth} { 
  | >{\centering\arraybackslash}X 
  | >{\centering\arraybackslash}X 
  | >{\centering\arraybackslash}X 
  | >{\centering\arraybackslash}X | }
 \hline
 Algorithm & std error $x$ & std error $y$ & std error $\theta$ (rad) \\
 \hline
%  NDT Actual & 1.01 & 51.30 & 0.02\\  % no outliars
NDT Actual & 0.554 & 0.490 & 0.00251 \\ %including outliars
\hline
ICET Actual & 0.1047 & 0.0545 & 0.00035\\
%\hline
%ICET Actual (bottom 90\%) & 0.0814 & 0.0429 & 0.00027\\
\hline
ICET Predicted & 0.101 & 0.0602 & 0.00035\\
\hline
\end{tabularx}

 \subsection*{Straight Tunnel Case (High Geometric Ambiguity)}
 
\begin{tabularx}{1.0\textwidth} { 
  | >{\centering\arraybackslash}X 
  | >{\centering\arraybackslash}X 
  | >{\centering\arraybackslash}X 
  | >{\centering\arraybackslash}X | }
 \hline
 Algorithm & std error $x$ & std error $y$ & std error $\theta$ (rad) \\
 \hline
%  NDT Actual & 1.01 & 51.30 & 0.02\\  % no outliars
NDT Actual & 0.330 & 39.26 & 0.00280 \\ %including outliars
\hline
ICET Actual & 0.0437 & N/A & 0.00031\\
%\hline
%ICET Actual (bottom 90\%) & 0.0343 & N/A & 0.00024\\
\hline
ICET Predicted & 0.0454 &  Excluded from Solution & 0.00032\\
\hline
\end{tabularx}

\vspace{3mm}

The results in the table indicate, at least in this simple test case, that ICET predicts accuracy well and that the algorithm identifies instances of geometric ambiguity, thereby verifying our two major claims about the ICET algorithm. It is interesting, too, that the accuracy of ICET, at least on this data set, is noticeably higher than the accuracy of NDT.

While ICET effectively negates the effects of ambiguity from extended surfaces (e.g. walls), ambiguity may still occur from other geometric conditions. Regularly spaced features of equal size (e.g. columns) may contribute to aliasing. However, extended flat surfaces remain the most common form of geometric ambiguity in automotive applications \parencite{Park}.

The ability for ICET to output a predicted error covariance makes the algorithm particularly useful in the context of Bayesian filtering algorithms, such as the extended Kalman filter. Constructing a Bayesian filter requires modeling sensor noise. Currently error models are available for individual lidar data points \parencite{Bonnabel} but relatively little information is available \parencite{kanhere2019lidar} to characterize the output of a lidar algorithm such as NDT. The ICET algorithm sidesteps this gap in the literature by directly estimating the accuracy of its output states. Moreover, the ICET accuracy prediction automatically adapts to reflect different environments and terrains, so it is not necessary to rely on the simplistic (though convenient) assumption that algorithm accuracy is the same under all conditions.

One limitation of ICET is that performance is somewhat dependent on the resolution and alignment of the voxel grid. Rectangular voxels are an attractive choice in that their regularity makes it easy to index and assign lidar points to individual voxels; however, rectangular voxels also have practical limitations, particularly in cases in which extended world features are not aligned with the grid. This issue can be resolved for NDT by using an overlapping grid \parencite{Hong}. For ICET, a new problem is that our dimension reduction approach assumes the extended object cuts through the middle of a cell rather than, say, across its corner. The apparent width of a wall passing through the corner of a voxel would appear to be small, and the feature would not necessarily be excluded if the wall were aligned at an arbitrary angle relative to the grid. This is expected to somewhat reduce the benefits of ICET's dimension-reduction step when applied to arbitrary data sets.

\section{Summary}
% <<1. State big picture of what paper is about. 2. Restate major contributions. 3. Briefly summarize simulation-based verification of those contributions. 4. State why this work is useful ... what will be enabled in the future by your research? >>

A new technique for matching point clouds, the Iterative Closest Ellipsoidal Transform (ICET), was introduced which provides two novel improvements over existing voxelized scan-match methods such as the Normal Distributions Transform (NDT). The first contribution of ICET is the ability to estimate solution-error covariance (and thus the expected standard deviation of error for each component of translation and rotation in the solution vector) as a function of scene geometry. This is accomplished by first subdividing each scan into voxel grids and treating the distribution of points in each voxel as a probability density function. A linear least-squares solution is then calculated to determine the optimal transformation to minimize the differences between distribution centers for the new scan and a reference scan. This process is iterated multiple times to account for nonlinearities. The second major contribution of ICET allows the algorithm to ignore contributions to the solution vector from locally ambiguous directions (such as along the surface of a long smooth wall) while keeping information in useful directions (such normal to the wall).

Performance of ICET was verified in a simulated 2D environment against a benchmark implementation of NDT. In Monte-Carlo simulations of 1000 trials, ICET achieved an 80\% reduction in estimation error while successfully predicting the standard deviation of error for all components of rotation and translation. The ability to estimate accuracy has utility for sensor fusion algorithms (e.g. a Kalman Filter), especially since ICET can dynamically adjust accuracy estimates on the fly, which avoids reliance on static - and likely overconservative - accuracy estimates.
Furthermore, in a pathological straight-tunnel simulation, ICET was able to successfully identify the ambiguous dimension in all trials and suppress the solution along that axis. 
By contrast, in the same situation, NDT naively provides inaccurate estimates without warning that an ambiguity is present. The ambiguity-handling properties of ICET are especially important for autonomous vehicle applications, where tunnels and urban canyons may introduce ambiguity in the form of the aperture problem.

\section*{acknowledgements}
The authors wish to acknowledge and thank the U.S. Department of Transportation Joint Program Office (ITS JPO) and the Office of the Assistant Secretary for Research and Technology (OST-R) for sponsorship of this work. Opinions discussed here are those of the authors and do not necessarily represent those of the DOT or other affiliated agencies. 

\section*{Appendix}
As discussed in Section \ref{AmbiguitySec}, the dimension-reduction process can, in rare cases, result in a solution singularity. These special cases can be detected numerically by analyzing (\ref{eq:condense3}) to compute the matrix \textit{condition number}, which for this symmetric and positive-definite matrix is the ratio of its maximum to minimum eigenvalue.  Matrices with high condition number are not practical to invert, and so in the case where the condition number is high (larger than, say, $10^{5}$), then the lowest eigenvalue of (\ref{eq:condense3}) can be eliminated iteratively until the condition number falls below the threshold. This process eliminates dimensions from the \textit{solution}, in contrast with the prior dimension-reduction step, which eliminated dimensions from the \textit{measurement}.

In order to reduce the dimension of the solution, start by defining an eigendecomposition for (\ref{eq:condense3}), involving the eigenvector matrix $\mathbf{V}$ and the eigenvalue matrix $\mathbf{\Gamma}$.

\begin{equation}\label{eq:eigen2}
    \mathbf{H}^T\mathbf{WH} = \mathbf{V}  \mathbf{\Gamma} \mathbf{V}^T
\end{equation}

Again the eigenvalues can be partitioned into a submatrix of preserved eigenvalues $\mathbf{\Gamma}_P$, which satisfy the condition number requirement, and a submatrix of eliminated eigenvalues $\mathbf{\Gamma}_N$, which includes the approximately singular eigenvalues, meaning the low eigenvalues removed to meet the condition number requirement. The eigenvector matrix $\mathbf{V}$ can also be partitioned into the preserved and eliminated eigenvectors,  $\mathbf{V}_P$ and $\mathbf{V}_N$, respectively. We can then solve a modified version of (\ref{eq:dxCorrection}) in the well-conditioned subspace. To this end, we define coordinates $\delta\mathbf{z}$ for the well-condition subspace where

\begin{equation}\label{eq:dx2dz} 
    \delta\mathbf{z} = \mathbf{V}_P^T \ \delta\mathbf{x}.
\end{equation}

Substituting (\ref{eq:eigen2}) into into (\ref{eq:dxCorrection}), multiplying by $\mathbf{V}_P^T$, and introducing (\ref{eq:dx2dz}), the iterative Newton-Raphson correction (\ref{eq:dxCorrection}) becomes

\begin{equation}\label{eq:dzCorrection}
    \delta \mathbf{z} = \mathbf{\Gamma}_P^{-1}\  \mathbf{V}_P^T \ (\mathbf{H}^T\mathbf{W} \Delta \mathbf{y})
\end{equation}

The corrections are accumulated into the variable $\mathbf{x}'$ (initialized to zero), such that

\begin{equation}\label{eq:replacement2}
    \mathbf{x}' \xrightarrow[]{} \mathbf{x}' + \mathbf{V}_P \ \delta \mathbf{z}.
\end{equation}

At each Newton-Raphson stage, the estimated state $\hat{\mathbf{x}}$ is computed as the initial state $\bar{\mathbf{x}}$ plus the correction from (\ref{eq:replacement2}).

\begin{equation}\label{eq:update2}
    \hat{\mathbf{x}} = \bar{\mathbf{x}} + \mathbf{x}'
\end{equation}

When the iterations converge, there is only confidence in the solution in the coordinates of the well-conditioned subspace,
associated with the directions of the vectors $\mathbf{V}_P$. For this reason, it is helpful to map the correction term $\mathbf{x}'$ back into the well-conditioned subspace as $\mathbf{z}$, in order to accurately express the solution covariance. This mapping has the form of a general observation equation, characterized by an observation matrix $\mathbf{C}$.

\begin{equation}
    \mathbf{z} = \mathbf{C} \mathbf{x}'
\end{equation}

In this case, the observation matrix is simply the projection matrix $\mathbf{C} = \mathbf{V}_P^T$.  The uncertainty associated with this the difference between the estimate $\hat{\mathbf{z}}$ and the true value $\mathbf{z_{true}}$ is characterized by the following covariance.

\begin{equation}\label{eq:reducedCov}
    E[(\hat{\mathbf{z}}-\mathbf{z}_{true})(\hat{\mathbf{z}}-\mathbf{z}_{true})^T] = \mathbf{\Gamma}^{-1}_P    
\end{equation}

Result (\ref{eq:reducedCov}) has a dimension that is smaller than the size of the full-state covariance (\ref{eq:Pmatrix}), reflecting the fact that there is no confidence in the solution in the eliminated directions, those associated with $\mathbf{V}_N$.

%\newpage
% the IEEE bibliography style matches the ION bibliography style guidelines.
% \bibliographystyle{IEEEtran}
% \bibliography{template_bibliography}
\printbibliography

\end{document}